# AutoScore-Survival: Developing interpretable machine learning-based time-to-event scores with right-censored survival data

Feng Xie[1,2], Yilin Ning[2], Han Yuan[2], Benjamin Alan Goldstein[1,3], Marcus Eng Hock Ong[1,4], Nan Liu[1,2,5,6]*, Bibhas Chakraborty[1,2,3,7]*

[1] Programme in Health Services and Systems Research, Duke-NUS Medical School, Singapore, Singapore
[2] Centre for Quantitative Medicine, Duke-NUS Medical School, Singapore, Singapore
[3] Department of Biostatistics and Bioinformatics, Duke University, Durham, NC, United States
[4] Department of Emergency Medicine, Singapore General Hospital, Singapore, Singapore
[5] Institute of Data Science, National University of Singapore, Singapore, Singapore
[6] Health Services Research Centre, Singapore Health Services, Singapore, Singapore
[7] Department of Statistics and Applied Probability, National University of Singapore, Singapore, Singapore

* Correspondence: liu.nan@duke-nus.edu.sg; bibhas.chakraborty@duke-nus.edu.sg

## Abstract

**Background:** Scoring systems are highly interpretable and widely used to evaluate time-to-event outcomes in healthcare research. However, existing time-to-event scores are predominantly created ad-hoc using a few manually selected variables based on clinician's knowledge, suggesting an unmet need for a robust and efficient generic score-generating method.

**Methods:** AutoScore was previously developed as an interpretable machine learning score generator, integrated both machine learning and point-based scores in the strong discriminability and accessibility. We have further extended it to time-to-event data and developed AutoScore-Survival, for automatically generating time-to-event scores with right-censored survival data. Random survival forest provides an efficient solution for selecting variables, and Cox regression was used for score weighting. We implemented our proposed method as an R package. We illustrated our method in a real-life study of 90-day mortality of patients in intensive care units and compared its performance with survival models (i.e., Cox) and the random survival forest.

**Results:** The AutoScore-Survival-derived scoring model was more parsimonious than survival models built using traditional variable selection methods (e.g., penalized likelihood approach and stepwise variable selection), and its performance was comparable to survival models using the same set of variables. Although AutoScore-Survival achieved a comparable integrated area under the curve of 0.782 (95% CI: 0.767-0.794), the integer-valued time-to-event scores generated are favorable in clinical applications because they are easier to compute and interpret.

**Conclusions:** Our proposed AutoScore-Survival provides an automated, robust and easy-to-use machine learning-based clinical score generator to studies of time-to-event outcomes. It provides a systematic guideline to facilitate the future development of time-to-event scores for clinical applications.

**Key Messages:**
- We developed AutoScore-Survival, an interpretable machine learning score generator, by extending AutoScore to survival data for automatically generating time-to-event scores with right-censored survival data
- We applied AutoScore-Survival to a real-life dataset from the electronic health record with 44,918 individual critical care admission episodes and developed a time-to-event score for predicting a patient's future survival within 90 days after admission.
- AutoScore-Survival-derived scores showed superiority in interpretability and accessibility as a parsimonious point-based integer score, compared with Cox regression and "black box" random survival forest.

## Introduction

The interpretable predictive model is essential for supporting medical decision-making, where doctors can easily understand how the models make predictions in a transparent manner. There has been a growth in inherently interpretable machine learning models[1,2], where risk scoring systems were highly preferred in healthcare settings. Recently, Ustun et al. developed Risk-calibrated Supersparse Linear Integer Model (RiskSLIM)[3] and further improved it through the optimization of risk scores[4]. Besides, we previously provided a practical solution, AutoScore, as an interpretable machine learning-based automatic clinical score generator. Users can automatically generate a data-driven clinical score given a dataset in various clinical applications, facilitating automated machine learning (AutoML)[5] solutions in healthcare. However, those models were initially designed for binary outcomes, and extending them to time-to-event outcomes is of great value.

There are different regression and machine learning options for the prediction of time-to-event outcomes. Typically these models generate a probability of not having the event (i.e., surviving) at a specified time point (e.g., predicting 30-day mortality). However, like most predictive models, such algorithms fail to generate an indicative score for straightforward risk stratification. In comparison, scoring systems would be strongly preferred in healthcare since they are highly transparent and interpretable, based on addition, subtraction, and multiplication of a few sparse numbers, facilitating clinical practice even without the need for a computer. At present, this type of time-to-event score has been pervasively used in healthcare, such as survival prediction score[6], Palliative Prognostic Score[7,8], Respiratory ECMO Survival Prediction[9] across different clinical disciplines. They were developed to support treatment decisions by forecasting the time to patient outcome (e.g., death or disease progression) or by projecting the change in risk over time. However, these time-to-event scores were created ad-hoc via manual variable selection based on expert opinion, suggesting the unmet need for a robust and efficient generic method for deriving time-to-event scores.

Traditionally, survival data are analyzed using Cox regression, where variables selection is predominantly performed by stepwise selection ( Akaike information criterion [AIC][10, 11] and the Bayesian information criterion [BIC][12, 13]) or by penalizing the partial likelihood[14] (i.e., least absolute shrinkage and selection operator [LASSO][15]). However, However, such approaches are not efficient when working with high-dimensional data, e.g., the electronic health records (EHR)[16]. Machine learning, such as random survival forest [17, 18] XGBoosting[19, 20] supporter vector machine (SVM)[21], and deep learning models (artificial neural network)[22] have been applied for more efficiently handling high-dimensional survival data, but most of them are black boxes that are challenging to comprehend. Thus, there is an unmet need to develop a parsimonious survival prediction model with easy access to validation in the context of high-dimensional EHRs.

To address these challenges, we extended previously mentioned AutoScore[23] to survival data and systematically presented AutoScore-Survival, a generic method for developing parsimonious time-to-event scores. The proposed AutoScore-Survival framework can automatically generate a single indicative score for predicting patients' time-to-event outcomes and was demonstrated to build an actual score for survival prediction of ICU patients. We also compared the AutoScore-Survival-created scores with other standard baselines.

## Methods

The AutoScore framework was developed to generate prediction scores for binary outcomes[23]. It consists of six modules: Module 1 ranks variables using machine learning methods, Module 2 categorizes continuous variables to deal with nonlinearity and simplify interpretation, Module 3 derives scores from a subset of variables using the logistic regression, Module 4 selects the best number of variables through parsimony plot, Module 5 allows fine-tuning of cut-offs for categorizing continuous variables for preferable interpretation and Module 6 performs the final performance evaluation of the score. Our proposed AutoScore-Survival method extends this framework can be extended to time-to-event data by modifying relevant modules. Figure 1 illustrates the six-module framework of AutoScore-Survival, where the modules to be modified from AutoScore are highlighted in xx and elaborated in detail in the following subsections.

**Variable ranking with random survival forests at Module 1**
In real-world clinical applications, we set up training, validation, and test set accordingly. The training set is used to derive the score. The validation set is used for intermediate performance evaluation and parameter selection. The test set acts as an unseen dataset and is used to generate the final model performance. Let $(t_i, \delta_i, X^i)$ denote the survival data for the $ith$ individual in the training set. $t_i$ denotes the time of the event if censoring indicator $\delta_i = 1$ and time of censoring if $\delta_i = 0$. $X^i$ denotes the vector of $p$ available variables. Our goal is to rank all $p$ available variables and select $m$ parsimonious variables ($m < p$) for the following score derivation. For simplicity of notation, we will omit $i$ in the subscript and superscript when no confusion arises.

We use random survival forest (RSF)[17, 18, 24], an ensemble machine learning algorithm, to analyze survival data and rank variables. It consists of a number of binary survival trees grown by a recursive splitting of tree nodes[25]. Central elements of the RSF algorithm are growing survival trees by maximizing survival difference (log-rank test statistic)[26] and estimating the survival probabilities based on the ensemble cumulative hazard function. RSF exerts two forms of randomization at the ensembling process: a bootstrapping sample of data and a randomly selected subset of variables. Averaging over survival trees, as well as the two forms of randomization, makes RSF much more accurate in prediction[20] and variable ranking[27, 28]. The details of RSF are elaborated in eTextbox 1 of the Appendix.

Then, variable importance is calculated based on the corresponding reduction of predictive accuracy when the variable of interest is replaced with its random permutation value[24]. Unlike the traditional Cox regression, RSF does not assume proportional hazard or any functional form for the hazard function and works well for high-dimensional EHR data[29]. The ranking list will be used in subsequent modules for model building.

**Score derivation by weighting and normalization at Module 3**
Similar to AutoScore, in AutoScore-Survival models are built using the training set by selecting top-ranking variables from the ranking list, and continuous variables are transformed into categorical variables. With the selected and transformed variables, we create a time-to-event score for the survival data based on Cox Regression[30], with which the points can be easily interpreted:

$$h(t,X) = h_0(t) \times e^{-(\beta_1 X_1 + \cdots + \beta_m X_m)} \tag{1}$$

where $t$ represents the survival time, $h(t,X)$ is the hazard function given variables $(X_1 \ldots X_m)$. $(\beta_1 \ldots \beta_m)$ are the coefficients for each variable, and $h_0(t)$ is the baseline hazard. The Cox regression does not make parametric assumptions on $h_0(t)$. Weibull and log-normal can also be used as the weighting function, where $h_0(t)$ is assumed specific functional forms [31].

Based on equation (1), a partial score is assigned to each category of the variable, which is derived from the coefficients through a two-step procedure. The first step is to change the reference category in each variable to the category with the smallest $\beta$ coefficient from the first-step regression such that all scores are non-negative. Next, the second-step regression is performed to generate new coefficients. The partial scores are derived from the second-step regression by dividing each coefficient by the minimum of all $\beta$'s, and the results are rounded to the nearest integer. With a partial integer score associated with each category of a variable, the total score for each patient is computed by summing up all partial scores.

**Model selection under the intermediate performance evaluation at Module 4**

The validation set is used for the intermediate performance evaluation. We use a survival parsimony plot to visualize the change in model performance with an increasing number of variables, which helps us select a model that balances prediction accuracy and parsimony. For time-to-event outcomes, the time-dependent area under the curve or AUC(t)[32] is applied to measure model performance, which is an extension of the commonly used area under the curve (AUC) for measuring predictive accuracy of a score when studying binary outcomes. We chose the AUC(t) defined by cumulative sensitivity and dynamic specificity (C/D) as recommended by a comprehensive review[33], as this definition has more clinical relevance and commonly been used by clinical application[34]. This AUC(t) is introduced as a function of time, estimated through the Kaplan-Meier estimator of survival function[32], to characterize how well the score can distinguish between subjects who had an event <= t and remained event-free at time t. To obtain a single overall performance metric in the parsimony plot, we derived the integrated AUC (iAUC), a weighted average of AUC(t)[35] over a follow-up period (i.e., from Day 1 to Day 90), summarizing the overall discrimination ability of the time-to-event score (see eTextbox 2 in Appendix for detail).

**Final predictive performance evaluation at Module 6**

We evaluate the final time-to-event score in the test set using multiple performance metrics. In addition to iAUC and AUC(t), we used the Harrell's concordance index (C-index)[36, 37], which is the proportion of concordant pairs (i.e., when the observation with a longer survival time has a larger time-to-event score) in all pairs formed in the test set. Thus, the C-index is able to summarize risk, event occurrence, and survival time in a single number to distinguish between well-behaved scores and quasi-random ones[38].

**Algorithm Implementation and empirical validations**

We implemented the AutoScore-Survival framework as an R package[39]. Given a new dataset with time-to-event outcomes and baseline covariates, the AutoScore-Survival package provides a pipeline of functions to split data and implement the six modules to generate the final scores that require minimal manipulation from users.

We demonstrated our AutoScore-Survival algorithm using the same dataset as our previous paper[23], including 44,918 ICU admissions from Beth Israel Deaconess Medical Center (BIDMC)[40] with 24 available variables of demographic information, vital signs, and lab tests at baseline (t=0). The survival status and the date of death were additionally obtained from the database to derive the 90-day survival as the primary outcome. The baseline characteristics of the dataset were described through univariable and multivariable Cox regression. The Kaplan-Meier survival curves were generated for different risk groups stratified by the scores and compared through the log-rank test. Furthermore, we compare the predicted and actual survival probabilities at different time points. To evaluate the performance of AutoScore-Survival, we compared it with several standard time-to-event prediction models. We considered the Cox model with (i) all variables and that with variables selected using (ii) stepwise and (iii) LASSO[41] approaches. The stepwise variable selection used AIC and considered both directions in each step. The regularization rate of LASSO was optimized through 10-fold cross-validation). We also built an RSF using all variables, with the widely-accepted default parameters[42] (i.e., 500 trees grown). Model performance was reported on the test set, and 100 bootstrapped samples were applied to calculate 95% confidence intervals (CI)[43].

## Results

**Cohort formation and basic covariates analysis**
Overall survival probability was estimated by the Kaplan-Meier method (see Appendix eFigure 1). 37,496 (83.5%) admission episodes survived longer than 90 days and were censored at the end of the 90-day observation window. 7722 (16.5%) episodes died within 90 days, with a median survival time of 15 (IQR: 6-38) days and a mean survival time of 24.7 days (SD=24.0). Table 1 summarises the univariable and multivariable Cox analyses of all prognostic factors. All variables except gender got $P < 0.001$, making it hard to select a parsimonious model according to $P$ values.

**Parsimony plot and time-to-event scores**
AutoScore-Survival selected seven variables by the parsimony plot (Figure 2a) based on the validation set, as it achieved a good balance between model performance (i.e., iAUC) and complexity (number of variables, $m$). When more variables were added to the time-to-event score, the performance was not markedly improved.

The seven-variable survival scores, derived from age, anion gaps, respiration rate, creatinine, temperature, blood urea nitrogen, and lactate levels, are tabulated in Table 2. The final score ranges from 0 to 100, where a smaller score indicates a higher survival probability. Table 3 shows different score intervals and their corresponding percentile survival time and survival probability estimated using the Kaplan-Meier method. The survival probability at 3, 7, 30, and 90 days decreases with increasing time-to-event scores, as expected. Scores larger than 60 correspond to a 90-day survival probability of lower than 50%. Table 2 and Figure 3(a) offer a correspondence of scores and predicted survival probability based on the training set. As shown in Figure 3(b), the time-to-event score is able to accurately stratify patients in the test set into risk groups based on the Kaplan-Meier Curve ($P < 0.0001$).

**Performance evaluation and comparison**
The performance of various methods evaluated in the unseen test set was reported in Table 4. The seven-variable AutoScore-Survival scores achieved an iAUC of 0.782 (95% CI: 0.767-0.794) and a C-index of 0.753 (95% CI: 0.74-0.762), comparable to the Cox regression with all 24 variables with an iAUC of 0.785 (95% CI: 0.768-0.798) and a C-index of 0.759 (95% CI: 0.748-0.769). LASSO and stepwise Cox regression also achieved a comparable iAUC of 0.782 (95% CI: 0.766-0.795) and 0.785 (95% CI: 0.772-0.799) as well. But both selected 17 or 22 variables, respectively, failing to filter out redundant information efficiently to build up a parsimonious model for easy interpretation, compared with only seven variables of the AutoScore-Survival. Although the full RSF model achieved the highest iAUC and C-index in our experiment, consisting of a number of separate survival trees makes it become a black box and not interpretable enough for real-world application. In terms of time-dependent

AUC(t=3,7,30,90), our seven-variable time-to-event score achieved comparable performances to 24-variable and stepwise Cox regression or LASSO.

## Discussion

In the present study, we developed AutoScore-Survival by extending the AutoScore method [23] to time-to-event outcomes and demonstrated its application by creating a time-to-event score using real-world data on 90-day survival in ICU. The score generated by the AutoScore-Survival was comparable with other standard survival prediction methods (i.e., Cox regression, LASSO, stepwise selection approach) in terms of discriminative capability. Although RSF outperformed at the model accuracy, more importantly, the AutoScore-Survival scores showed superior interpretability as a parsimonious point-based single indicative score for predicting patients' overall survival. This study's novelty is to integrate the advantages of RSF for robust variable selection and Cox regression in its accessibility for a generic methodology of quickly creating parsimonious time-to-event scores based on survival data. Future studies could apply it to various real-world clinical data to develop more useful time-to-event scores across different clinical settings.

The proposed AutoScore-Survival has several advantages in generating time-to-event outcome predictive scores. First, AutoScore-Survival could generate a parsimonious score by RSF-based variable selection, which has been shown to identify critical variables in high-dimensional data with multicollinearity[44]. In our example, AutoScore-Survival achieved comparable performance with survival models of all 24 variables by selecting only seven variables, while LASSO and stepwise approaches failed to select a parsimonious list of variables. Second, AutoScore-Survival categorizes continuous variables to account for possible non-linear (e.g., U-shaped[45]) effects. Although there are some advanced regression methods to handle nonlinearity[46], categorization is favorable in clinical and epidemiological applications for straightforward indication of identifying high-risk and low-risk values. Third, the score-based model is much easier to use and understand for healthcare professionals. Compared with other decimal predictors or probability outputs, integer time-to-event scores let users make quick predictions by simple arithmetic and gauge the effect of changing variables[38], even without the need for complex calculation. Thus, our scores have the advantage of accessibility and easy implementation, especially at the bedside. Furthermore, we derived scores from the Cox regression that is familiar to clinicians and does not make restrictive assumptions on the baseline hazard. Still, our R package[39] allows users to choose parametric survival models (Weibull regression and log-normal) as the weighting function. Besides right-censored data, left-censored and interval-censored Cox models[47] could be further extended by creating a different survival object using the *Surv* function.

We illustrated the application of AutoScore-Survival in acute care settings, where the better performance at the early time (e.g., the higher AUC(t=3) and AUC(t=7)) could help identify patients who need intensive care or extra medical attention. AutoScore-Survival is also useful for clinical decision-making on chronic diseases, e.g., when on cancer treatment and management[48]. For example, a small score or long survival time might indicate more aggressive and progressive cancer treatment, and a short survival time might be the indicator for palliative care[49] to optimize the life quality and mitigate patients' suffering. In our study, Table 2 and Figure 3(a) linked the time-to-event scores with survival time and probability, where it can help stratify patients into risk groups for appropriate allocation of therapeutic and care strategies[7].

There has been a growing interest in developing time-to-event scores in the clinical literature. For example, Kim et al.[50] recently built a scoring system to predict the overall survival of patients with advanced gastric cancer. Becker et al.[51] and Sharma et al.[52] also developed a time-to-event score for patients with general cancer and hepatocellular carcinoma, respectively. However, all of them were developed in an ad-hoc way. AutoScore-Survival provides a systematic guideline for automated development of time-to-event scores, contributing to data-driven research on various types of diseases.

This study also has several limitations. First, the dataset used in this study was based on 24 variables available from the EHR, where some other relevant variables were not available. The primary goal of this empirical implementation is to demonstrate the effectiveness of AutoScore-Survival for generating time-to-event scoring models. Second, the scores represent the relative risk in the population if baseline hazard is ignored under Cox regression and adjustment is needed in different horizons. Third, this is the initial development of AutoScore-Survival, where we selected commonly used methods to build our framework and demonstrated its usage using EHR data. Further development should extend the framework with advanced algorithms.

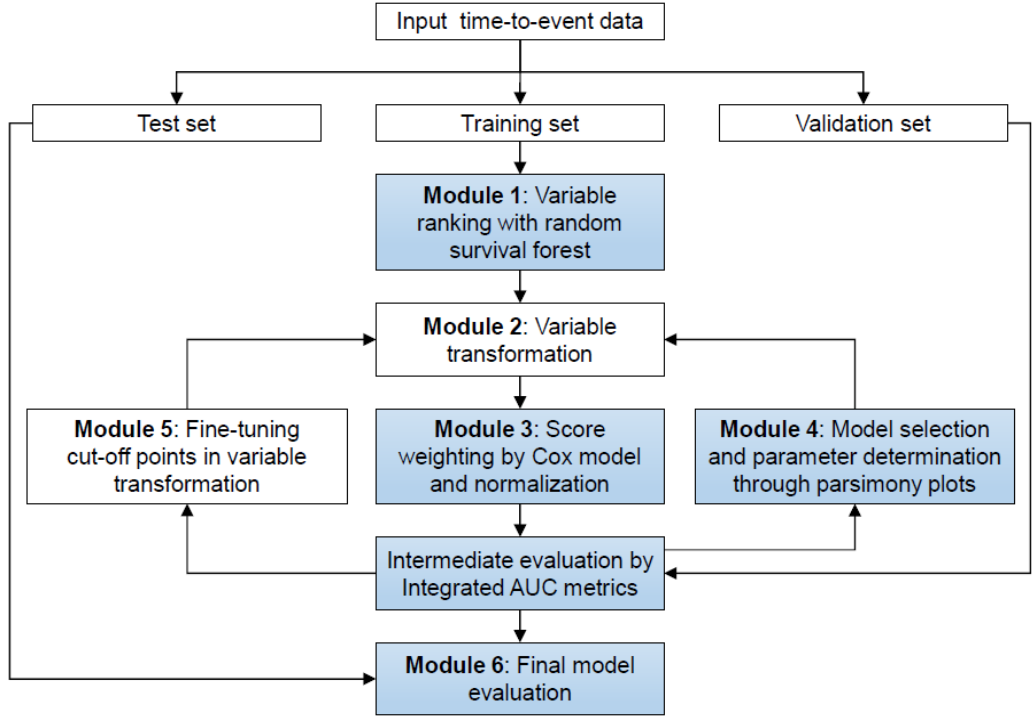

Figure 1. Flowchart of the AutoScore-Survival framework. The blue shadow blocks are unique in the AutoScore-Survival compared with original AutoScore

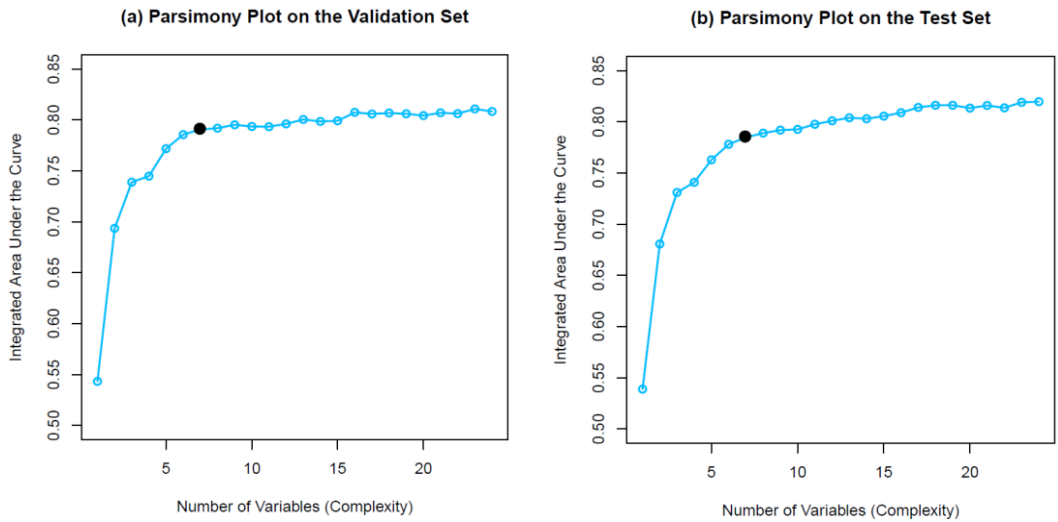

Figure 2. Parsimony plot (model performance versus complexity) by the integrated area under the curve on (a) validation set and (b) the test set. The solid black dots show the selected point for achieving the model parsimony (i.e., number of variables *m* = 7)

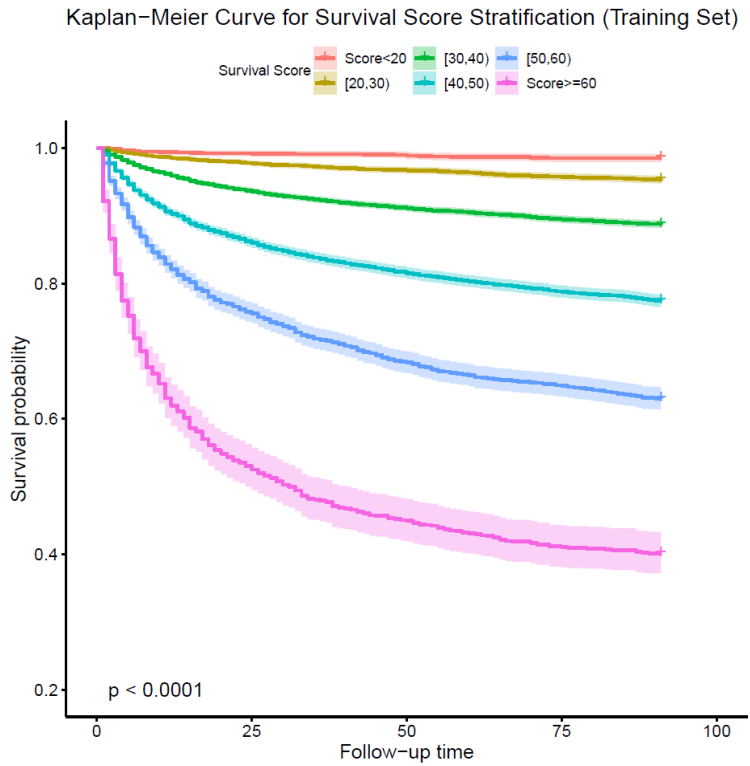

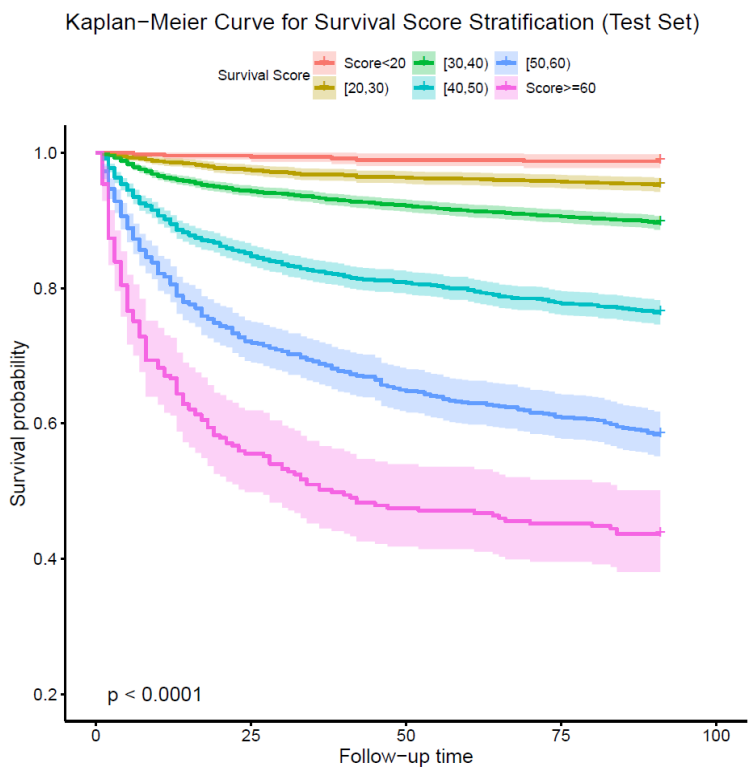

Figure 3. Actual overall survival through risk stratification by AutoScore-Survival scores on the (a) training set and (b) test set (Kaplan-Meier estimates)

Table 1: Univariable and multivariable survival analysis of all variables in the study cohort (N=44,918)

|  | Unadjusted HR (95% CI) | p-Value | Adjusted HR (95% CI) | Adjusted p-Value |
|---|---|---|---|---|
| **Age (years)** | 1.032(1.031-1.034) | <0.001 | 1.027(1.025-1.029) | <0.001 |
| **Gender** | | | | |
| Female | Baseline | | Baseline | |
| Male | 0.966(0.922-1.011) | 0.135 | 1.088(1.037-1.142) | 0.001 |
| **Ethnicity** | | | | |
| White | Baseline | | Baseline | |
| Hispanic | 0.368(0.305-0.444) | <0.001 | 0.844(0.695-1.026) | 0.089 |
| Asian | 0.527(0.480-0.578) | <0.001 | 0.968(0.873-1.073) | 0.533 |
| African | 0.482(0.456-0.510) | <0.001 | 0.891(0.834-0.953) | 0.001 |
| Others | 0.502(0.386-0.653) | <0.001 | 1.369(1.045-1.795) | 0.023 |
| **Insurance** | | | | |
| Medicare | Baseline | | Baseline | |
| Government | 1.432(1.166-1.757) | 0.001 | 1.146(0.933-1.407) | 0.194 |
| Medicaid | 2.717(2.253-3.276) | <0.001 | 1.184(0.975-1.439) | 0.089 |
| Private | 1.311(1.082-1.587) | 0.006 | 1.056(0.870-1.280) | 0.582 |
| Self Pay | 1.363(0.989-1.879) | 0.058 | 1.622(1.176-2.237) | 0.003 |
| **Heart rate (beats/min)** | 1.017(1.015-1.018) | <0.001 | 1.016(1.014-1.018) | <0.001 |
| **Systolic blood pressure (mmHg)** | 0.986(0.985-0.988) | <0.001 | 0.988(0.986-0.991) | <0.001 |
| **Diastolic blood pressure (mmHg)** | 0.978(0.976-0.98) | <0.001 | 0.990(0.985-0.995) | <0.001 |
| **Mean arterial pressure(MAP; mmHg)** | 0.977(0.975-0.979) | <0.001 | 1.015(1.009-1.021) | <0.001 |
| **Rrespiration rate (breaths/min)** | 1.097(1.092-1.103) | <0.001 | 1.058(1.052-1.065) | <0.001 |
| **Temperature(°C)** | 0.693(0.666-0.721) | <0.001 | 0.796(0.764-0.829) | <0.001 |
| **Peripheral capillary oxygen saturation (SpO2; % )** | 0.923(0.916-0.931) | <0.001 | 0.981(0.972-0.991) | <0.001 |
| **Glucose (mg/dL)** | 1.003(1.003-1.004) | <0.001 | 1(1-1.001) | 0.577 |
| **Anion gap (mEq/L)** | 1.108(1.102-1.114) | <0.001 | 1.036(1.023-1.05) | <0.001 |
| **Bicarbonate (mmol/L)** | 0.961(0.956-0.966) | <0.001 | 0.984(0.973-0.995) | 0.005 |
| **Creatinine (μmol/L)** | 1.095(1.084-1.106) | <0.001 | 0.911(0.893-0.93) | <0.001 |
| **Chloride (mEq/L)** | 0.970(0.966-0.974) | <0.001 | 0.962(0.953-0.972) | <0.001 |
| **Lactate (mmol/L)** | 1.243(1.229-1.257) | <0.001 | 1.117(1.1-1.134) | <0.001 |
| **Hemoglobin (g/dL)** | 0.849(0.839-0.86) | <0.001 | 0.690(0.657-0.724) | <0.001 |
| hematocrit (%) | 0.956(0.952-0.96) | <0.001 | 1.101(1.082-1.119) | <0.001 |
| **Platelet (thousand per microliter)** | 1(1-1) | 0.014 | 0.999(0.999-0.999) | <0.001 |
| **Potassium (mmol/L)** | 1.129(1.09-1.169) | <0.001 | 0.886(0.851-0.922) | <0.001 |
| **Blood urea nitrogen (BUN; mg/dL)** | 1.017(1.016-1.018) | <0.001 | 1.013(1.012-1.014) | <0.001 |
| **Sodium (mmol/L)** | 0.990(0.985-0.996) | <0.001 | 1.020(1.009-1.031) | 0.001 |
| **White blood cells (thousand per microliter)** | 1.007(1.006-1.007) | <0.001 | 1.005(1.004-1.006) | <0.001 |

Table 2. Seven-variable AutoScore-Survival-derived scoring model

|  | Variable and Interval | Partial Score |
|---|---|---|
| Age (years) | <30 | 0 |
|  | [30,48) | 8 |
|  | [48,78) | 15 |
|  | [78,85) | 22 |
|  | >=85 | 25 |
| Blood urea nitrogen (BUN; mg/dL) | <7.5 | 0 |
|  | [7.5,8.25) | 17 |
|  | [8.25,12) | 1 |
|  | >=12 | 8 |
| Respiration rate (breaths/min) | <12 | 6 |
|  | [12,16) | 0 |
|  | [16,22) | 4 |
|  | >=22 | 11 |
| Creatinine (mg/dL) | <0.5 | 14 |
|  | [0.5,0.8) | 4 |
|  | [0.8,1.6) | 0 |
|  | >=1.6 | 1 |
| Anion Gap (mEq/L) | <15 | 0 |
|  | [15,20) | 4 |
|  | >=20 | 7 |
| Lactate (mmol/L) | <1 | 0 |
|  | [1,2.5) | 3 |
|  | [2.5,4) | 6 |
|  | >=4 | 15 |
| Temperature (°C) | <36 | 11 |
|  | [36,36.5) | 4 |
|  | [36.5,37.3) | 0 |
|  | [37.3,38) | 3 |
|  | >=38 | 6 |

Table 3: Time-to-event score intervals and their corresponding percentile survival time or survival probability at different time points

| Score Value | Percent of patients | 10th Percentile Survival Time (days) | 25th Percentile Survival Time (days) | Median Survival Time (days) | Survival probability at three days (%) | Survival probability at seven days (%) | Survival probability at 30 days (%) | Survival probability at 90 days (%) |
|---|---|---|---|---|---|---|---|---|
| ≤20 | 5.33% | 90+ | 90+ | 90+ | 100.0% | 99.8% | 99.4% | 98.7% |
| (20, 30] | 18.46% | 90+ | 90+ | 90+ | 99.5% | 99.2% | 97.1% | 95.4% |
| (30, 40] | 39.55% | 87 | 90+ | 90+ | 99.2% | 97.6% | 93.9% | 89.8% |
| (40, 50] | 24.16% | 12 | 90+ | 90+ | 96.3% | 92.5% | 83.5% | 76.6% |
| (50, 60] | 9.60% | 5 | 19 | 90+ | 92.8% | 85.6% | 70.6% | 58.6% |
| >60 | 2.91% | 2 | 7 | 38 | 83.9% | 72.8% | 53.3% | 43.7% |

Table 4: Performance of the AutoScore-Survival and other baseline models.

|  | AutoScore-Survival | Full Cox Regression | Full Random Survival Forest | Regularized Cox regression (LASSO) | Stepwise Cox regression |
|---|---|---|---|---|---|
| m | 7 | 24 | 24 | 17 | 22 |
| iAUC | 0.782 (0.767-0.794) | 0.785 (0.768-0.798) | 0.843 (0.829-0.854) | 0.782 (0.766-0.795) | 0.785 (0.772-0.799) |
| C-index | 0.753 (0.740-0.762) | 0.759 (0.748-0.769) | 0.808 (0.801-0.817) | 0.755 (0.746-0.766) | 0.759 (0.751-0.769) |
| AUC (t=3) | 0.805 (0.776-0.827) | 0.781 (0.750-0.810) | 0.852 (0.829-0.871) | 0.782 (0.754-0.817) | 0.782 (0.752-0.815) |
| AUC (t=7) | 0.787 (0.771-0.805) | 0.787 (0.768-0.804) | 0.843 (0.828-0.859) | 0.785 (0.762-0.804) | 0.788 (0.770-0.808) |
| AUC (t=30) | 0.773 (0.756-0.785) | 0.786 (0.774-0.800) | 0.841 (0.829-0.851) | 0.780 (0.767-0.795) | 0.785 (0.774-0.798) |
| AUC (t=90) | 0.763 (0.751-0.773) | 0.778 (0.766-0.788) | 0.829 (0.820-0.838) | 0.774 (0.764-0.785) | 0.778 (0.769-0.790) |

iAUC, the integrated AUC(t); AUC(t), the area under the ROC(t) curve

C-index, concordance index; LASSO, Least Absolute Shrinkage and Selection Operator for Cox regression

## Supplementary Materials/ Appendix

### eTextbox 1. Detailed demonstration of the Random Survival Forest

The general flow of forming a Random Survival Forest (RSF) [17]:

1. Draw bootstrap samples from the original data. Each bootstrap sample would exclude on average 37% of the data, denoted as out-of-bag (OOB) data for the following extraction of variable importance.
2. Grow a survival tree for each bootstrap sample. At each node of the tree, randomly select $r$ input variables, where $r = floor(\sqrt{m})$. The node is split using the $r$ input variables by log-rank splitting rules.
3. Calculate a cumulative hazard function (CHF) for each tree and weighted average them to obtain the ensemble CHF.

The detailed demonstration for CHF calculation is shown below[17].

Let $(t_{il}, \delta_{il})$ denote the survival data for $ith$ subject in the $lth$ terminal node of one survival tree. The ordered distinct event times in the $lth$ terminal node are denoted as $0 < t_{1l} < t_{2l} < ... < t_{cl} < \infty$, where $t_{sl}$ denotes the $sth$ ordered statistic among distinct observation and censoring time with $t_{0l} = 0$.

We denote $d_{sl}$ and $R_{sl}$ as the number of deaths and the subjects at risk at the time, respectively. Then the estimate for the CHF and survival function are given by:

$$H(t|X^i) = \sum_{t_{sl} \leq t} \frac{d_{sl}}{R_{sl}}, for\ X^i \in l$$

$$S(t|X^i) = e^{-H(t|X^i)}, for\ X^i \in l$$

The bootstrap ensemble CHF takes the average over the B survival tree:

$$H_{RSF}(t|X^i) = \frac{1}{B}\sum_{b=1}^{B} H_b(t|X^i)$$

Where $H_b(t|X^i)$ represents the CHF from a tree on $sth$ bootstrap sample

**eTextbox 2. Detailed demonstration of time-dependent AUC and iAUC**

ROC curves are a popular method for displaying the sensitivity and specificity of a score (continuous diagnostic marker) for binary outcomes. However, measuring discrimination in survival analysis is more complicated and ambiguous than that in logistic regression. Thus, the ROC curves were further extended to survival data, i.e., time-dependent ROC or ROC(t). They have different versions or definitions, and we used ROC(t) of cumulative sensitivity and dynamic specificity (C/D)[33] in our study.

Under the survival setting, ROC curves that vary as a function of time may be more appropriate. In this study, we used For example, regarding the $ith$ subject, the outcome $D_i(t) = 1$ if a patient has died prior to time $t$ and zero otherwise. Hence, at a given time $t$ and a cut-off value $c$, the sensitivity and specificity can be defined as below for the $ith$ subject

$$sensitivity: S_n(c,t) = P(score_i > c \mid D_i(t) = 1)$$
$$specificity: S_p(c,t) = P(score_i \leq c \mid D_i(t) = 0)$$

The time-dependent ROC curve at time $t$ is defined as:

$$ROC(c,t) = S_n[\{1 - S_p\}^{-1}(\psi, t), t]$$

Where $\{1 - S_p\}^{-1}(\psi, t) = inf\{c: 1 - S_p(c,t) \leq \psi\}$. In this definition, the ROC curve is cumulative or dynamics as all the events occurring before time $t$ are considered as positive cases.

The AUC statistics at time $t$ is defined as the area under the ROC curve at time $t$ [32]:

$$AUC(c,t) = \int ROC(c,t)dt$$

Time-dependent ROC analysis requires a continuous diagnostic marker (i.e., score) as the input. For other baseline models, we extracted the corresponding linear predictors (i.e. $\beta_1 X_1 + \cdots + \beta_m X_m$) from Cox-based baselines, predicted $log(T)$ from Weibull regression, and ensemble predicted values from the RSF for comparisons. The iAUC can be viewed as the global average of the AUC(t), represented as a weighted average of AUC(t) [35]:

$$iAUC = \int_0^T AUC(t) \cdot \omega(t)dt$$

It is estimated by a weighted sum of AUC (t) with weights $S(t-1) - S(t)$ from the KM estimator of the event time distribution over the follow-up time range $(0, T)$. We use estimators proposed by Heagerty and colleagues.

**eTextbox 3. Detailed demonstration for concordance index (C-index)**

Assume we have a sample of $n$ subjects. Let $t_1, t_2, \ldots, t_n$ denote the survival times and $p_1, p_2, \ldots, p_n$ denote the predicted probabilities of survival. Given that (at least one subject is not censored), we assign 1 to each concordant pair and 0 to each discordant pair[36].

$$c_{ij} = 1 \text{ if } p_i > p_j \text{ and } t_i < t_j \text{ or } p_i < p_j \text{ and } t_i > t_j,$$
$$c_{ij} = 0 \text{ otherwise}$$

Let $\Omega$ denote a set of all eligible pairs of subjects $(i, j)$ and $|\Omega|$ denote the number of all pairs. Then we have

$$C - index = \frac{1}{|\Omega|} \sum_{(i,j) \in \Omega} c_{ij}$$

**eFigure 1**. Survival outcomes estimated by the Kaplan-Meier method for the whole cohort (N=44,918)

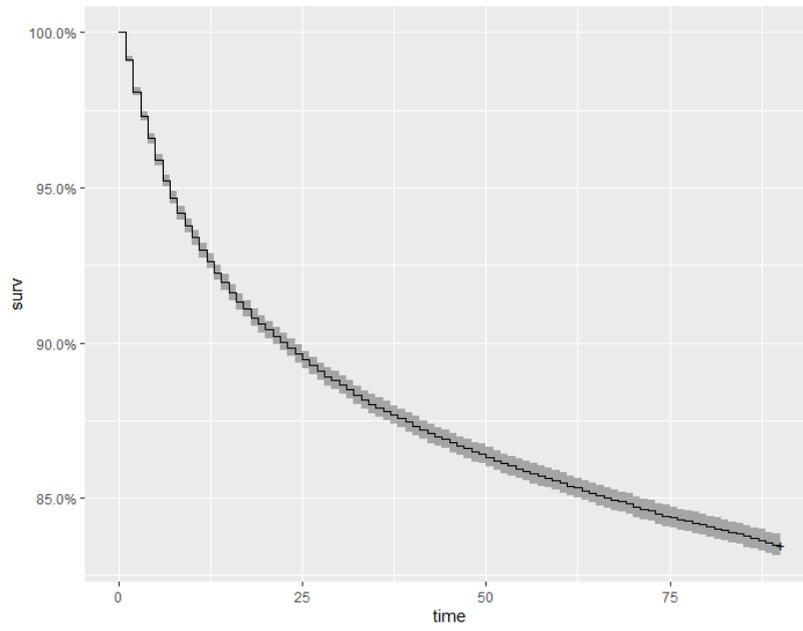